\documentclass{article}

\usepackage{arxiv}

\usepackage[utf8]{inputenc} % allow utf-8 input
\usepackage[T1]{fontenc}    % use 8-bit T1 fonts
\usepackage{hyperref}       % hyperlinks
\usepackage{url}            % simple URL typesetting
\usepackage{booktabs}       % professional-quality tables
\usepackage{amsfonts}       % blackboard math symbols
\usepackage{nicefrac}       % compact symbols for 1/2, etc.
\usepackage{microtype}      % microtypography
\usepackage{lipsum}
\usepackage{graphicx}

\usepackage{times}
\usepackage{latexsym}
\usepackage{float}

\usepackage[T1]{fontenc}

\usepackage[utf8]{inputenc}

\usepackage{microtype}

\author{
	Mahdi Khodabandeh\\
	Department of Computer Engineering\\
	University of Guilan\\
	Rasht, Guilan, Iran\\
	\texttt{20.mahdikh.0@gmail.com} \\
	\And
	Ghazal Shabani\\
	Department of Computer Engineering\\
	University of Guilan\\
	Rasht, Guilan, Iran\\
	\texttt{ghazalshabani.gca@gmail.com} \\
	\And
	Arash Yousefi Jordehi\\
	Department of Computer Engineering\\
	University of Guilan\\
	Rasht, Guilan, Iran\\
	\texttt{arashy76@phd.guilan.ac.ir} \\
	\And
	Seyed Abolghasem Mirroshandel\\
	Department of Computer Engineering\\
	University of Guilan\\
	Rasht, Guilan, Iran\\
	\texttt{mirroshandel@guilan.ac.ir} \\
}

\usepackage{amsmath,amsfonts}
\usepackage{array}
\usepackage[caption=false,font=normalsize,labelfont=sf,textfont=sf]{subfig}
\usepackage{textcomp}
\usepackage{stfloats}
\usepackage{url}
\usepackage{verbatim}
\usepackage{booktabs}

\usepackage{latexsym}

\usepackage[T1]{fontenc}
\usepackage[utf8]{inputenc}

\usepackage{microtype}

\usepackage{multirow}
\usepackage{amsthm}
\usepackage{csquotes}
\usepackage{graphicx}
\usepackage[acronym,nomain]{glossaries}
\usepackage{tabularray}

\usepackage{inconsolata}
\usepackage{xcolor}
\usepackage{soul}
\usepackage{amssymb}
\usepackage[colorinlistoftodos,textsize=tiny]{todonotes}
\usepackage{arydshln}
\usepackage{enumitem} % For customizing lists

\usepackage{algorithm}
\usepackage{algpseudocode}  % better than algorithmic, more powerful

\usepackage{rotating}
\usepackage{needspace}

\usepackage{pifont}
\usepackage{multirow}
\usepackage{booktabs}
\usepackage{tabularx}
\usepackage{tabularray}
\UseTblrLibrary{diagbox, booktabs} % For diagbox within tblr
\usepackage{csquotes}
\usepackage[normalem]{ulem}
\usepackage{arydshln}

\usepackage{algpseudocode}

\usepackage{minted}
\usepackage{mdframed}

\surroundwithmdframed[
backgroundcolor=white,  % Background color is white
linecolor=white,        % Border color is black
]{minted}

\setminted[text]{
	fontsize=\small,
	breaklines,
	style=default,         % Code color is black
}

\usepackage[acronym,nomain]{glossaries}

\newacronym{nlp}{NLP}{Natural Language Processing}
\newacronym{dl}{DL}{Deep Learning}
\newacronym{bert}{BERT}{Bidirectional Encoder Representations from Transformers}
\newacronym{t5}{T5}{Text-to-Text Transfer Transformer}
\newacronym{mpqa}{MPQA}{Multi-Perspective Question Answering}
\newacronym{ml}{ML}{Machine Learning}
\newacronym{rl}{RL}{Reinforcement Learning}
\newacronym{nncp}{NNCP}{Neural Network Compression}
\newacronym{ese}{ESE}{Expressive Subjective Element}
\newacronym{ds}{DS}{Direct Subjective}
\newacronym{ose}{OSE}{Objective Speech Event}
\newacronym{nltk}{NLTK}{Natural Language Toolkit}
\newacronym{lstm}{LSTM}{Long Short-Term Memory}
\newacronym{gru}{GRU}{Gated Recurrent Unit}
\newacronym{cnn}{CNN}{Convolutional Neural Network}
\newacronym{rnn}{RNN}{Recurrent Neural Network}
\newacronym{llrd}{LLRD}{Layer-wise Learning Rate Decay}
\newacronym{mtl}{MTL}{Multi-Task Learning}
\newacronym{json}{JSON}{JavaScript Object Notation}
\newacronym{html}{HTML}{HyperText Markup Language}
\newacronym{xml}{XML}{eXtensible Markup Language}
\newacronym{sota}{SOTA}{state-of-the-art}
\newacronym{fsl}{FSL}{Few-Shot Learning}
\newacronym{fs}{FS}{Few-Shot}
\newacronym{al}{AL}{Active Learning}
\newacronym{llm}{LLM}{Large Language Models}
\newacronym{ir}{IR}{Intermediate Representation}
\newacronym{gelu}{GeLU}{Gaussian Error Linear Unit}
\newacronym{relu}{ReLU}{Rectified Linear Unit}
\newacronym{a2c}{A2C}{Advantage Actor-Critic}

\theoremstyle{definition}

\definecolor{arashcolor}{rgb}{0.5, 0.8, 0.5} % Light green
\definecolor{seyedcolor}{rgb}{0.5, 0.5, 0.8} % Light blue
\definecolor{ghazalcolor}{rgb}{0.8, 0.5, 0.5} % Light red
\definecolor{mahdicolor}{rgb}{0.8, 0.8, 0.5} % Light yellow

\usepackage[colorinlistoftodos,textsize=tiny]{todonotes}

\title{Seq2Seq2Seq: Lossless Data Compression \\via Discrete Latent Transformers \\and Reinforcement Learning}

\begin{document}
	\maketitle
	\begin{abstract}
		Efficient lossless compression is essential for minimizing storage costs and transmission overhead while preserving data integrity. Traditional compression techniques, such as dictionary-based and statistical methods, often struggle to optimally exploit the structure and redundancy in complex data formats. Recent advancements in deep learning have opened new avenues for compression; however, many existing approaches depend on dense vector representations that obscure the underlying token structure. To address these limitations, we propose a novel lossless compression method that leverages Reinforcement Learning applied to a T5 language model architecture. This approach enables the compression of data into sequences of tokens rather than traditional vector representations. Unlike auto-encoders, which typically encode information into continuous latent spaces, our method preserves the token-based structure, aligning more closely with the original data format. This preservation allows for higher compression ratios while maintaining semantic integrity. By training the model using an off-policy Reinforcement Learning algorithm, we optimize sequence length to minimize redundancy and enhance compression efficiency. Our method introduces an efficient and adaptive data compression system built upon advanced Reinforcement Learning techniques, functioning independently of external grammatical or world knowledge. This approach shows significant improvements in compression ratios compared to conventional methods. By leveraging the latent information within language models, our system effectively compresses data without requiring explicit content understanding, paving the way for more robust and practical compression solutions across various applications.
		
	\end{abstract}
	%%
	
	%% main text
	\section{Introduction} \label{sec:intro}
	% \Arashnote{Problem Statement: Introduce the challenge of data compression and the potential role of machine learning.}
	%\Seyed{Add some sentences here to define the problem and its importance. Then, next sentences are good.}
	Data compression refers to the technique of transforming data into a format that occupies less storage space than the original~\cite{salomon2002data}, all while maintaining an acceptable level of accuracy. This method is recognized as one of the key approaches to information encoding. In the context of networking, it is often termed bit-rate reduction. Notably, Morse Code, introduced in 1848, is regarded as the earliest form of modern data compression~\cite{lelewer1987data, mahoney2013data, graham1994signal}.
	Most lossless compression techniques utilize statistical models and mathematical approaches to compress data. Additionally, multimedia techniques leverage specific features of the original file to eliminate redundant data~\cite{9418257}. Similarly, we can take advantage of patterns in data to achieve compression.
	
	% \Arashnote{Significance of T5: Discuss why T5 is a suitable choice for this task, focusing on its capabilities in sequence modeling and text generation.}
	The \gls*{t5} language model~\cite{raffel2020exploring} is recognized as one of the top-performing architectures in \gls*{nlp} benchmarks, featuring an encoder-decoder transformer design. It is pre-trained on extensive datasets, including the C4 dataset, as well as task-specific datasets such as bug fixing and code summarization, which provides a strong foundation for understanding and generating data~\cite{mastropaolo2021studyingusagetexttotexttransfer}. The model handles inputs of different lengths and can be easily scaled up by adjusting the number of parameters, ranging from 60 million to 11 billion, with multiple configurations available for the model~\cite{article}.
	
	% \Arashnote{Objective: Clearly state the goal of the paper, such as improving compression rates using a novel approach.}
	
	% \Arashnote{Contributions: List the main contributions, such as a novel idea or algorithm and integration of T5 with RL. Use a numbered list.}
	
	%\replaced[id=ghazal]
	{The primary objective of this paper is to develop and evaluate a novel, lightweight lossless data compression framework that combines the pattern recognition capabilities of \glspl*{llm}~\cite{mirtaheri2025comparative} with the adaptive decision-making of \gls*{rl} techniques, while remaining computationally tractable for execution on typical personal computers. Unlike resource-intensive neural compression approaches, we present an efficient framework based on the \gls*{t5} architecture that operates entirely in the discrete token space. This token-based design enables effective compression while maintaining low memory and processing requirements, making the system practical for deployment on consumer-grade hardware without specialized acceleration.}
	%{The primary objective of this paper is to introduce and evaluate a novel lossless data compression method that integrates the capabilities of \glspl*{llm} with \gls*{rl} techniques. Specifically, we propose a compression framework based on the \gls*{t5} architecture, where data is encoded into sequences of discrete tokens rather than traditional dense vector representations.}
	By leveraging an off-policy \gls*{rl} algorithm, our approach optimizes sequence length to minimize redundancy while preserving semantic integrity, prioritizing practicality and accessibility over achieving the highest possible compression ratios. Compression strategies typically require customization based on the specific characteristics of tasks or datasets~\cite{shandilya2024tacorltaskawareprompt}. Generalizing \gls*{rl} policies across tasks or compression rates without retraining remains a challenge due to the conflicts between the exploration limitations of \gls*{rl} and diverse nature of data distributions~\cite{Yarats2021ReinforcementLW}. However, our approach is designed to be broadly applicable, demonstrating strong generalization across different datasets and compression scenarios without the need for task-specific fine-tuning. Another key consideration in RL-based compression algorithms is the trade-off between multiple objectives such as reducing model size, maintaining accuracy~\cite{wang2019haqhardwareawareautomatedquantization}, and meeting latency or energy constraints~\cite{10181715}. This multi-objective optimization is complex and requires carefully designed reward functions~\cite{9878432}. This method departs from conventional auto-encoder-based techniques by maintaining a token-based structure that better aligns with the original data format, enabling higher compression ratios without relying on predefined grammatical rules or external world knowledge. The achieved compression ratio may not surpass state-of-the-art deep learning-based compression models, but the ability to run efficiently on personal computers makes this approach a viable and scalable solution for real-world applications where resource constraints are a concern.
	
	%\section*{Contributions of the Paper}
	In the realm of deep learning-based compression, this research offers a novel perspective by combining the strengths of \gls*{llm} and deep reinforcement learning. We make the following key contributions:
	
	\begin{itemize}
		\item Development of a compression technique that dynamically optimizes compression strategies.
		\item Integration of the \gls*{t5} model with \gls*{rl}.
		\item Designing a model that compresses data through discrete \gls*{ir}.
		\item A scalable framework that is comparable to state-of-the-art machine learning-based compression techniques.
		\item Utilization of \glspl*{llm} to leverage contextual features for compression.
		\item A framework that ensures efficient execution without requiring specialized computational resources.
	\end{itemize}
	
	To promote reproducibility and facilitate further research, we plan to release all source code publicly upon acceptance of this paper. Furthermore, we intend to provide a user-friendly implementation of our framework, such as a well-documented API, to encourage its broader adoption.
	%\Seyed{Add a paragraph to describe the content of the next sections of the paper!}
	
	The remainder of this paper is organized as follows: Section 2 reviews previous work, classifying lossless compression methods into learning-based and conventional strategies. Section 3 introduces fundamental concepts related to \gls*{rl} and the \gls*{t5} model, providing the necessary theoretical background. Section 4 details our methodology, including the design of the RL-based compression system, training procedures, and optimization techniques. Section 5 presents experimental results, including ablation studies and performance comparisons. Finally, Section 6 highlights key findings, discusses limitations, and offers recommendations for future research.
	
	\section{Literature Review}
	
	% \Arashnote{Review existing literature on data compression techniques, focusing on those using machine learning and reinforcement learning. I write specific aspects in highlighted text in this section.}
	
	%\subsection{Data Compression Techniques}
	%\Seyed{This paragraph can be used as the first paragraph of the introduction section. Here, you should include 3-4 sentences reviewing the literature, emphasizing that you have categorized existing methods into two types. Additionally, I removed the title of Section 2.1, 'Data Compression Techniques,' as it is unnecessary. I also changed the next two sub-subsections to subsections.!} 
	
	Lossless compression has been extensively studied, with existing methods broadly categorized into classic approaches and machine learning-based techniques. Classic methods, such as Huffman coding and Lempel-Ziv~\cite{Zeeh2003TheLZ} rely on statistical and dictionary-based encoding to achieve efficient compression. However, machine learning-based methods leverage data-driven approaches to enhance compression efficiency. By navigating this evolution map, we have explored \gls*{rl} techniques as a promising direction for optimizing compression strategies.
	
	\subsection{Classic Methods}
	
	LZ77 is a dictionary-based lossless compression algorithm that utilizes a dynamically sliding window. This window consists of two primary components: the search buffer (historical dictionary) on the left and the look-ahead buffer on the right. The search buffer stores previously seen sequences of symbols, while the look-ahead buffer contains upcoming symbols yet to be encoded. As the window slides through the input stream, the algorithm attempts to find the longest matching sequence in the search buffer that corresponds to the beginning of the look-ahead buffer. Once a match is found, a token is generated, consisting of the match length, the offset to the start of the match in the dictionary, and the match itself. This technique enables efficient compression by eliminating redundant data while maintaining a straightforward decoding process ~\cite{salomon2002data}. The LZ77 algorithm laid the foundation for several derivatives, including LZ78, Lempel-Ziv-Welch (LZW), and XZ that incorporates additional techniques like Markov chain modeling for better prediction of data patterns~\cite{doi:10.1137/S1052623496304700}. GZIP is another strong variation of LZ77 that encodes the identified repeated pattern with Huffman coding~\cite{10127236}.
	
	Another fundamental method in entropy coding is arithmetic coding, which differs significantly from traditional prefix-based coding schemes like Huffman coding. Instead of representing symbols as discrete codewords, arithmetic coding encodes the entire message as an interval within [0,1). The algorithm creates a probability model based on the frequency of each symbol in the message and dynamically divides the interval based on the model. With each new symbol processed, the interval is refined according to the assigned probability of that symbol, thereby reducing the range in which the final encoded value resides. Since frequently occurring symbols contribute to minimal reductions in interval size, the algorithm achieves near-optimal compression efficiency. Arithmetic coding is widely used in applications such as image compression and video encoding due to its superior efficiency in handling variable-length code representations~\cite{10.1145/214762.214771}.
	
	Range encoding is a variant of arithmetic coding that eliminates both contextual and alphabetic redundancies in digital messages while maintaining similar compression effectiveness. Like arithmetic coding, range coding represents a message as a progressively refined interval within a numerical range. Each symbol shrinks this interval based on its probability. However, unlike traditional arithmetic coding which operates on real numbers, range coding uses fixed-precision integer arithmetic. This approach is computationally simpler and more practical for software implementations since it avoids costly floating-point operations. Unlike traditional prefix-based methods, range encoding operates on a continuous range, enabling it to achieve compression ratios closer to the theoretical entropy limit~\cite{martin1979range}.
	
	\subsection{Machine Learning Based Methods}
	% \Arash{Please write this comprehensively based on the papers and books, and their references materials.}
	\gls*{nncp} is a lossless data compression algorithm that leverages sequence modeling techniques to model the probability distribution of the input data. The first version of \gls*{nncp} employs a \gls*{lstm}~\cite{hochreiter1997long} to detect sequential dependencies within the input, allowing it to predict the likelihood of upcoming symbols with high accuracy. The probability distribution is then encoded using arithmetic coding. This effective approach achieved compression rates that are competitive with traditional statistical and dictionary-based algorithms. Same trained neural network weights are utilized in the decompression process which makes \gls*{nncp} a lossless compression algorithm.
	The second version of \gls*{nncp} replaces the \gls*{lstm}  backbone with a Transformer-based model. This version is more efficient in encoding and decoding since the transformer architecture captures long-range dependencies and leverages parallelization benefits. Other modifications in this version include changing the activation functions from \gls*{relu} to \gls*{gelu} and utilizing learned relative positional embeddings instead of sinusoid relative positional embeddings. These refinements, alongside other architectural enhancements, resulted in improved compression ratios even for diverse datasets~\cite{bellard2021nncp}.
	
	CMIX is a lossless compression algorithm that, like NNCP, results in a high computational cost to improve compression ratio. The algorithm also employs ensemble learning to predict the probability of each bit in the input, utilizing a diverse set of models to handle various data types, such as text and images. The predictions of all independent models of the ensemble are then combined into a single probability using a technique called context mixing. In this technique, there is some sets of manually defined contexts. Each set has its own learning rate and just a small subset of neurons in the network activate due to the context-dependent activations. This adaptive process allows CMIX to adjust its predictions dynamically based on the nature of the data, making it particularly effective for datasets that contain a mix of text, images, and structured data. Once the final probability distribution is established, CMIX encodes the data using arithmetic coding. CMIX is a viable choice for domains where compression efficiency is prioritize over processing speed due to its adaptability across multiple data types. However it is generally impractical for real-time applications~\cite{knoll2014cmix}.
	
	Transformer-based lossless compression framework is another compression framework based on probabilistic prediction task. The methodology consists of three primary stages. In data preparation stage, a large-scale dataset of 165GB was compiled from raw byte sequences including text, images, and audio. No explicit feature engineering is used so that the transformer model can learn patterns purely from the data distribution. The core architecture follows a decoder-only transformer model which is particularly suited for causal modeling. Each token is predicted based on previous context without any knowledge of future. A probability distribution for each byte based on previous observations is the final output of training. These probabilities are then fed into arithmetic coding. This approach dynamically adjusts encoding based on learned probability distributions which enhances compression efficiency~\cite{heurtel2024compression}.
	
	\section{Background}
	{Before delving into the specifics of our method, it is essential to understand the foundational technologies it builds upon. This section outlines two core components: the Transformer-based \gls*{t5} architecture, which serves as the backbone of our model's language processing capabilities, and the reinforcement learning (RL) framework, which enables the model to make structured decisions in a sequential context. Together, these elements provide the necessary tools for learning efficient compression strategies under a unified learning framework.}
	
	\subsection{T5 Architecture}
	% \Arashnote{We should talk about the architecture of Transformers and go into the detail of T5 structure. Explain how T5 can be adapted for specific tasks, including compression.}
	
	The T5 language model is based on the original transformer architecture which means it contains both encoder and decoder, along with self-attention mechanism~\cite{10.5555/3295222.3295349}. It employs two types of attention masking: fully visible and causal masking. Attention masking forces the model to attend to specific tokens and ignore the others. The encoder utilizes fully visible masking, also known as bidirectional masking, which allows the self-attention mechanism to consider all tokens at each time-step~\cite{9621874}. Conversely, causal masking is applied in the decoding phase to prevent the model from attending to future tokens which is a crucial behavior for text generator models~\cite{haviv2022transformerlanguagemodelspositional}. For instance, at time-step 3, the model can only attend to input tokens 1 through 3.
	The \gls*{t5} model is trained on a diverse set of corpora to provide a uniform framework for solving text processing problems without requiring any architectural modifications. Task-specific prefixes are employed to guide the model in generating outputs that correspond to each specific task. For instance, an input sentence like \enquote{translate English to German: ML is fun!} is used for a translation task, where the prefix \enquote{translate English to German:} guides the model to generate the appropriate output~\cite{raffel2020exploring}.
	
	\subsection{Reinforcement Learning Basics}
	% \Arashnote{I think it is good to have several paragraphs describing the advantage of RL algorithms. Provide a brief overview of the actor-critic method and its application in sequential decision-making.}
	
	% \Arash{Please write an introduction about RL, and use this citation I put here.} \cite{sutton2018reinforcement}
	
	With the increasing popularity of \gls*{rl},
	and its ability to solve problems that other machine learning techniques fall short of solving, it has been widely adopted in fields such as robotics, finance, healthcare, and gaming, where decision-making under uncertainty is crucial.
	
	In \gls*{rl}, an \textbf{agent} learns to interact with an \textbf{environment} to maximize \textbf{cumulative reward}~\cite{aritonang2025hidden} by taking \textbf{actions} according to a \textbf{policy} $\pi$. A policy is a mapping from perceived \textbf{states} of the environment to actions taken in those states. At each step, the agent takes an action and receives the reward (a real value) from the environment. This number indicates how good or bad the action was~\cite{sutton2018reinforcement}.
	%\Arash{I suggest introducing symbols of policy, reward, etc. (Using Greek symbols for say) in these sections to familiarize the reader with terminologies and symbols in our pseudo-code and algorithms.}
	
	Since the compression task can be formulated as a sequential decision-making problem, \gls*{rl} appears to be an effective solution due to its capability to learn optimal strategies through interaction with dynamic environments. By receiving direct feedback, agents can learn from both successes and failures in their actions. This trial-and-error approach is crucial because selecting a token is an interdependent action with long-term consequences. Moreover, \gls*{rl} acts well in changing conditions and environments like compression where the context can vary significantly~\cite{Zhou2021AnAA}.
	
	We employ the \gls*{a2c} algorithm, a fundamental approach in \gls*{rl} that combines the strengths of both policy-based and value-based methods. In \gls*{a2c}, the actor selects actions based on the current policy, while the critic estimates the expected return from a given state to guide policy updates. To improve learning stability and efficiency, \gls*{a2c} uses an advantage function which quantifies how much better a chosen action is compared to the average action at a given state~\cite{Wang2022RecursiveLS}. This interaction between the actor and critic not only enhances sample efficiency but also ensures continuous policy refinement.
	
	\section{Methodology}
	The main challenge in using neural networks for compression lies in identifying an optimal \gls*{ir} (i.e., the compression format) that is both entropy-efficient and fast for compression and decompression. Typically, this is achieved through auto-encoders that generate fixed-dimension vectors. Although this representation is differentiable and can be easily optimized using gradient descent, it is inefficient in terms of data usage. Each FP16 value consumes 2 bytes, and a large number of embedding dimensions is often required. Moreover, these dimensions cannot dynamically scale with the input data, which is crucial for effective compression. 
	
	% \Arash{Entropy-efficiency might not be understandable for the reader. Did we declare it in previous parts? If not, you can explain it here or as a footnote maybe.} 
	%\Arash{If we have any references it would be good to cite them. If it's not, that is not a problem.}
	
	Entropy-efficient in this context refers to the encoding that uses a number of bits per token near the Shannon entropy limit~\cite{shannon1948mathematical}. For an input with symbols $x_i$ and probabilities $p(x_i)$, entropy is:
	
	\begin{equation}
		H(X) = -\sum_{i} p(x_i) \log p(x_i)
	\end{equation}
	
	The number represents the lower bound of the average number of bits needed to represent each token without loss of information~\cite{Park01011998}.
	
	Being entropy efficient requires a uniform distribution of \gls*{ir} for all possible sequences, which is the same as keeping the distribution of each token uniform of all possible tokens.
	The most entropy-efficient \gls*{ir} is also the optimal \gls*{ir} according to the training metric, and brings it down to 0.
	%\Arash{It would be better if you support your claims with references. Or you can write some mathemtical proofs.}
	
\begin{figure} [H]
		\centering
		\includegraphics[scale=0.5]{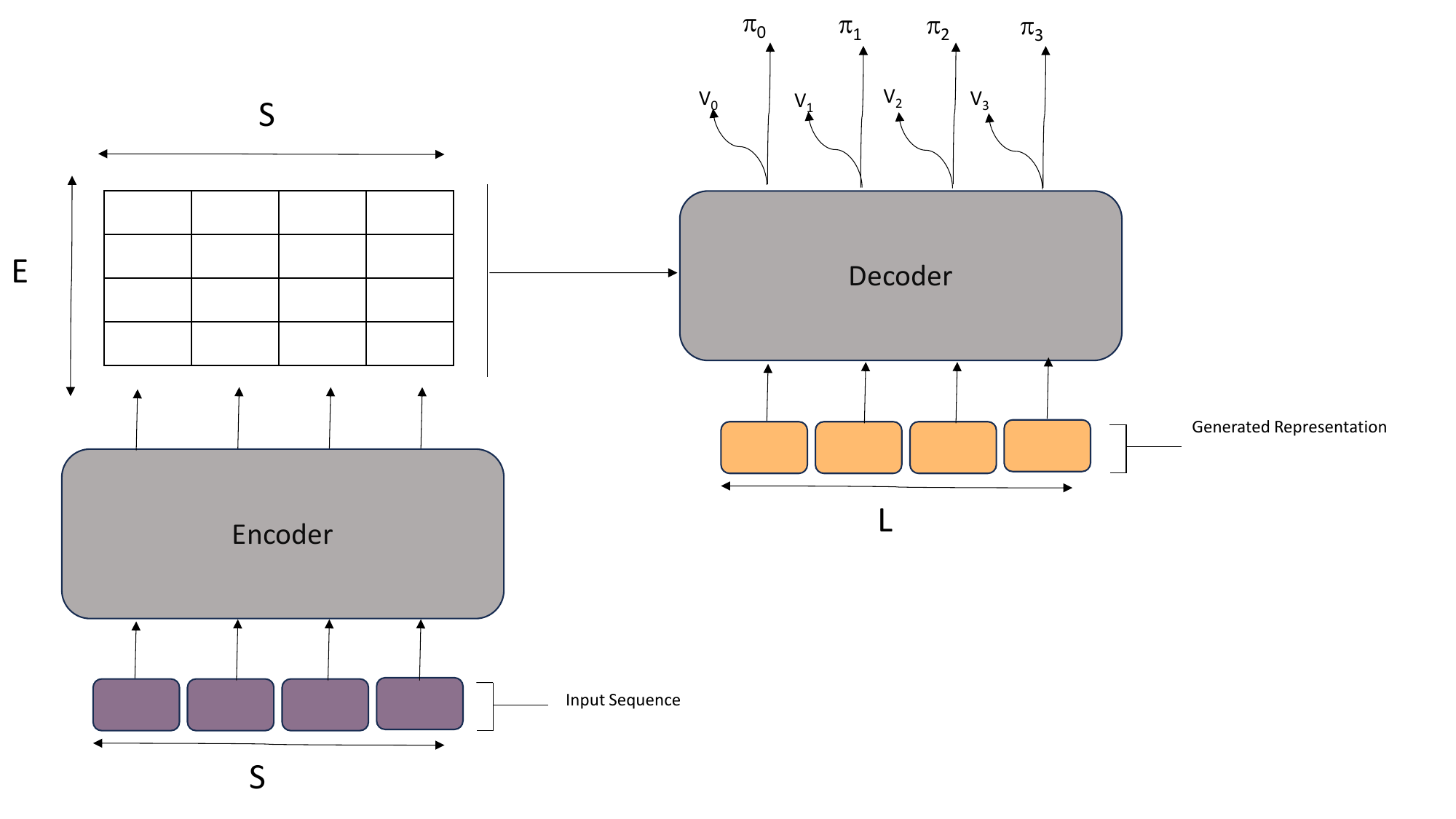}
		\caption{Structural overview of the compression network. The encoder processes an input sequence of length \( S \), generating embeddings of size \( S \times E \). The decoder reconstructs the original input from the compressed sequence of length \( L \), using both the generated representation and learned embeddings. The value head estimates compression quality, while the policy head predicts the next token.
		}
		\label{fig:encoder-decoder}
	\end{figure}

	We propose a sequence of tokens from a fixed alphabet, designed by the compression/decompression neural networks. This approach is more data-efficient, as discrete values are more effectively represented in computers using bits. Additionally, the sequence length is dynamic, allowing it to scale according to the entropy of the input data.
	The structural overview of the compression network is illustrated in Figure~\ref{fig:encoder-decoder}.
	
\begin{figure} [H]
		\centering
		\includegraphics[scale=0.45]{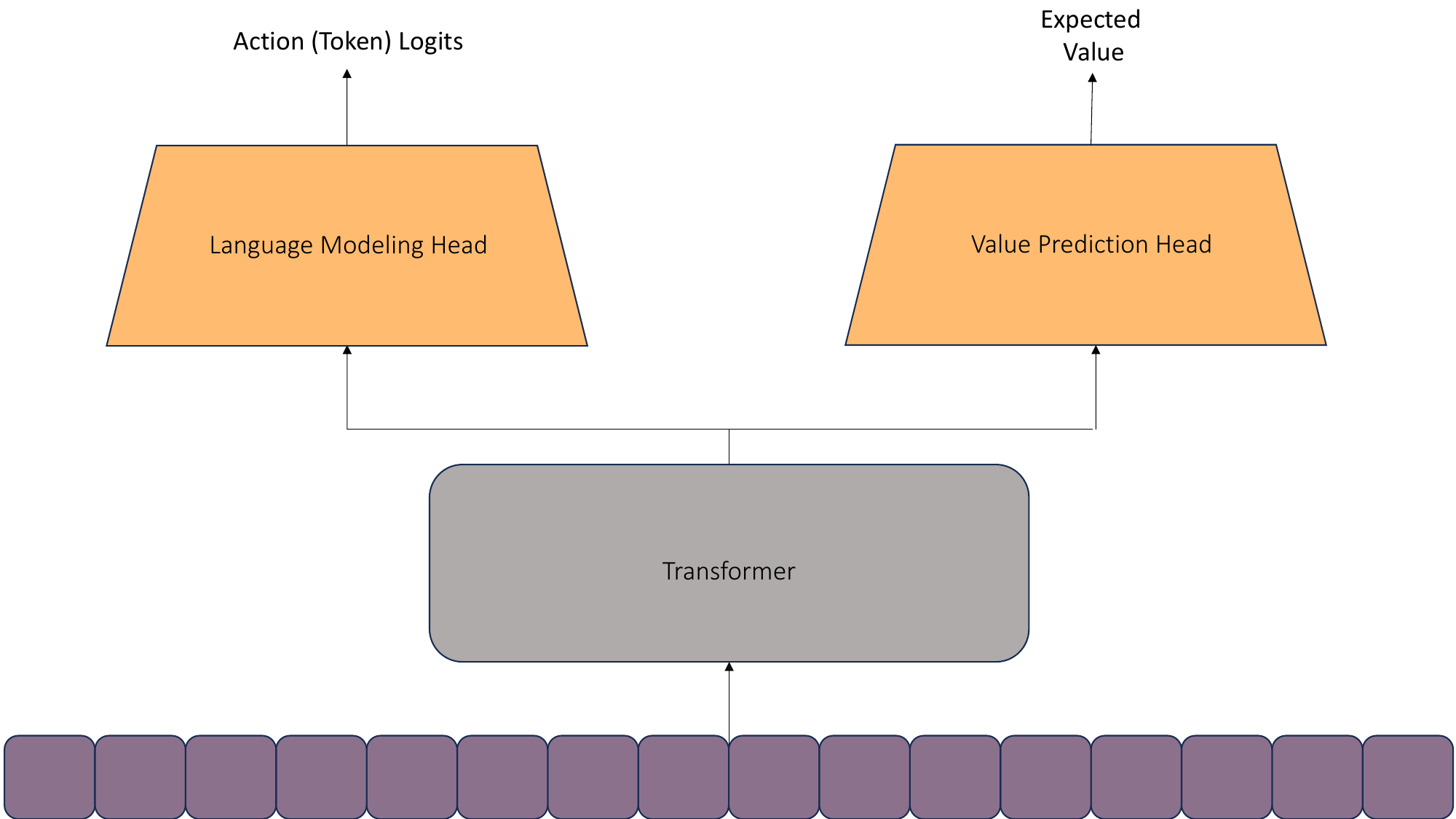}
		\caption{A2C implementation. The language modeling head is used as actor to choose tokens. The value head guides the actor by estimating value.
		}
		\label{fig:heads}
	\end{figure}

	% \Arash{I think it is better to contrast your novelty in design more and depict it as a figure specifically.}
	Generating a compressed format from the input is a decision problem that affects data loss in decompression, so we model it as an \gls*{rl} problem with token selection being the action, and a reward computed based on the metrics mentioned above. The compression network includes an extra pooling layer as a critic to predict values for \gls*{rl}. The interaction between the policy head and the value head in the A2C framework is shown in Figure~\ref{fig:heads}.

	Our methodology involves sampling a fixed-size chunk of input data and passing it through the compressor, which generates a compressed representation. This representation is not predetermined but rather learned dynamically. The decompressor is then trained to reconstruct the original data from this compressed form. To enforce an effective compression strategy, we define a reward function that guides the compressor’s learning process.
	
	Specifically, we define the reward function for each episode as follows:
	\begin{equation}
		r = - (|\overline{c}| + \mathcal{L}_D)
	\end{equation}
	where $\mathcal{L}_D$ represents the decompressor's loss when attempting to reconstruct the original input, $|\overline{c}|$ is a predefined cost factor     for the \gls*{ir}, and is proportional to the size of the compressed representation. This reward function effectively balances two competing objectives:
	
	\begin{itemize}
		\item Minimizing $\mathcal{L}_D$ ensures that the decompressor can accurately reconstruct the input from the compressed representation.
		\item The term $c \cdot n$ penalizes longer compressed representations, encouraging the compressor to generate more compact outputs.
	\end{itemize}
	
	The underlying motivation for this reward structure is to treat $\mathcal{L}_D$ as the direct measure of information loss incurred due to compression, while $c \cdot n$ acts as a regularization term that discourages unnecessary verbosity in the compressed representation. By optimizing for this reward, the model learns to strike a balance between preserving information and achieving higher compression ratios.
	
	This reinforcement learning framework, guided by the defined reward function, enables an adaptive compression mechanism that dynamically adjusts to varying input structures while maintaining efficient compression performance.
	
	% \Arashnote{
		% We require a two-column figure that illustrates the overall architecture of our framework, encompassing the input, model, and output components. Please include any elements you believe will enrich the reader's understanding and enhance the appeal of this work.}
	% \Arashnote{In three or more paragraphs, talk about the flow of the diagram, discuss each part in detail. For example, what is the input? Write equations for it. for output as well.}
	
	% \Arash{I suggest having another figure, which is kind of zoom-in on the connection of Transformer and A2C, that denotes what parts and elements of Transformers are utilized.}

	% \Arashnote{After presenting an overview of the system, please write a complete detailed pseudo-code of actor-critic algorithm (as Algorithm in Latex). It can be as long as you want. Then you write explanations for the algorithm line-by-line.}
	
	This algorithm trains two models:
	
	\begin{itemize}
		\item \textbf{Compressor (Encoder)}: A sequence-to-sequence (seq2seq) model trained using \gls*{rl}.
		\item \textbf{Decompressor (Decoder)}: A seq2seq model trained using standard conditional language modeling (LM).
	\end{itemize}
	
\begin{figure} [H]
		\centering
		\includegraphics[scale=0.45]{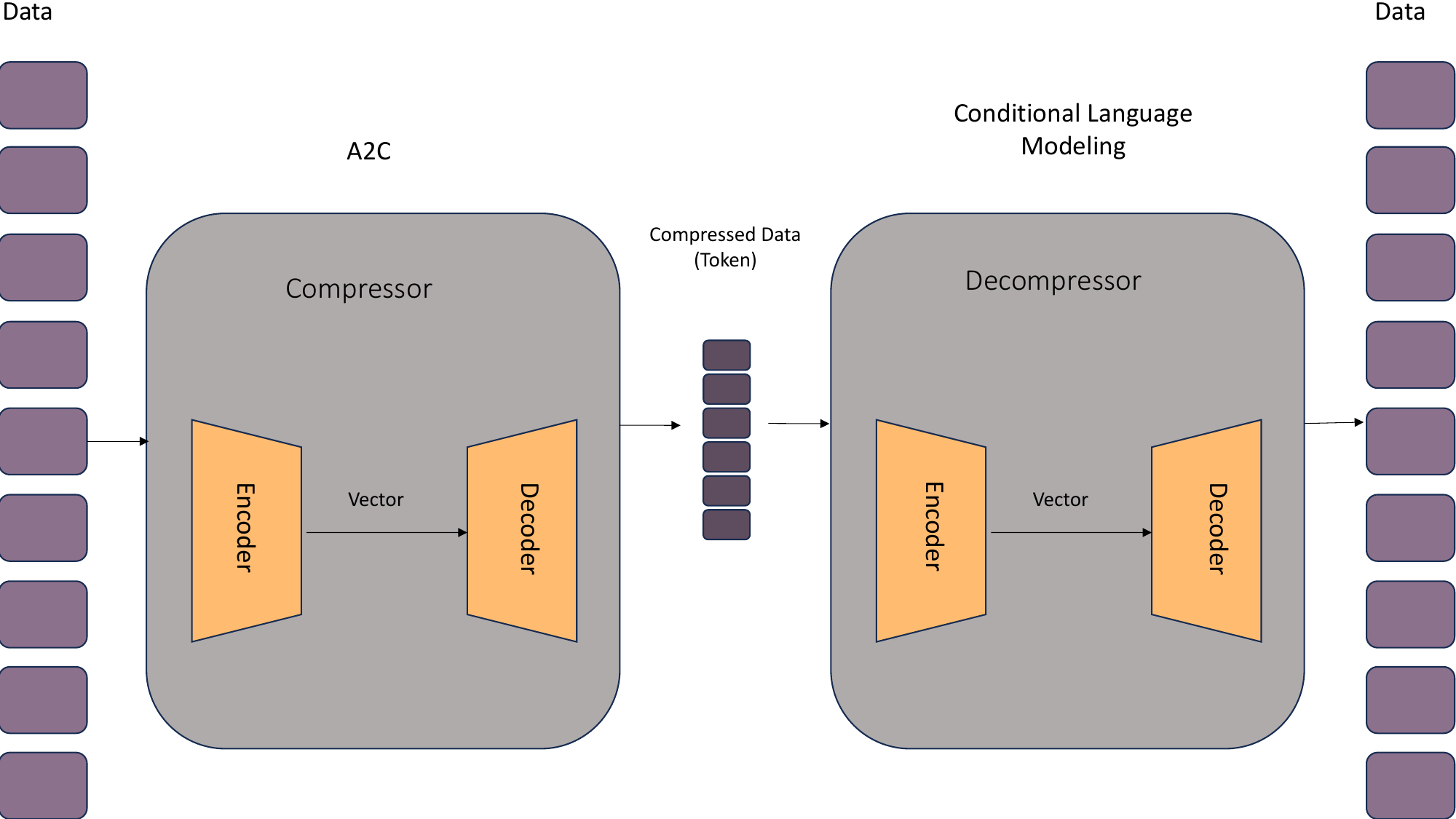}
		\caption{A conceptual view of our architecture. In this diagram the gray boxes represent Seq2Seq models which automatically generate a uniform token representation for both compressor and decompressor. The left box performs A2C algorithm to compress the input data and the right box tries to reproduce the original input from the compressed data.
		}
		\label{fig:latent-space}
	\end{figure}

	The compressor is trained to minimize the length of the compressed sequence while ensuring that the decompressor can accurately reconstruct the original sequence. The reward function for the compressor combines the compressed sequence length and the decompression loss. The following provides a detailed explanation of the algorithm. 
	A conceptual overview of the full architecture is presented in Figure~\ref{fig:latent-space}.
	
	\begin{algorithm}[H]
		\caption{Reinforcement-Based Sequence Compressor Training}
		\begin{algorithmic}[1]
			\Require Input batch $x \in \mathcal{X}$, tokenizer $\mathcal{T}$, compressor $\mathcal{C}$, decompressor $\mathcal{D}$
			\For{each batch $x$}
			\State $\text{input\_ids} \gets \mathcal{T}(x)$
			\State $\overline{c}, \hat{v} \gets \mathcal{C}(\text{input\_ids})$ \Comment{Compress input; estimate state value}
			\State $\overline{d} \gets \mathcal{D}(\overline{c})$ \Comment{Reconstruct original from compressed}
			\State $\mathcal{L}_D \gets \mathcal{L}_{\text{LM}}(\overline{d}, \text{input\_ids})$ \Comment{Language modeling loss}
			\State $\theta_{\mathcal{D}} \gets \theta_{\mathcal{D}} - \eta \nabla_{\theta_{\mathcal{D}}} \mathcal{L}_D$ \Comment{Update decompressor}
			\State $r \gets - (|\overline{c}| + \mathcal{L}_D)$ \Comment{Reward: compression length + reconstruction quality}
			\State $\mathcal{L}_{\text{Actor}} \gets \mathbb{E}[-\log \pi(\overline{c}) \cdot ((r + \gamma \hat{v'}) - \hat{v})]$
			\State $\mathcal{L}_{\text{Critic}} \gets \mathbb{E}[(r + \gamma \hat{v'} - \hat{v})^2]$
			\State $\theta_{\mathcal{C}} \gets \theta_{\mathcal{C}} - \eta \nabla_{\theta_{\mathcal{C}}} (\mathcal{L}_{\text{Actor}} + \mathcal{L}_{\text{Critic}})$
			\EndFor
		\end{algorithmic}
	\end{algorithm}

	As illustrated in Figure~\ref{fig:heads}, the compressor is implemented as an encoder-decoder transformer with two heads at the decoder output: a language modeling head, which acts as the actor by producing token logits, and a value head, which predicts a scalar reward estimate for each state.
	
	Given a full input sequence, compression is performed autoregressively to produce $\overline{c}$, the compressed representation. All subsequent steps—including decoding, loss computation, and policy/value updates—are performed in parallel over the full trajectory. The decompressor $\mathcal{D}$ reconstructs the original input from $\overline{c}$, and a language modeling loss $\mathcal{L}_D$ is computed between the predicted and original token sequences.
	
	The reward is defined as a negative combination of reconstruction loss and compressed sequence length:
	
	$$
	r = - (|\overline{c}| + \mathcal{L}_D)
	$$
	
	The actor loss uses the predicted advantage:
	
	$$
	\mathcal{L}_{\text{Actor}} = \mathbb{E}\left[-\log \pi(\overline{c}) \cdot \left(r + \gamma \hat{v}' - \hat{v} \right)\right]
	$$
	
	and the critic loss follows a standard TD error:
	
	$$
	\mathcal{L}_{\text{Critic}} = \mathbb{E}\left[\left(r + \gamma \hat{v}' - \hat{v} \right)^2\right]
	$$
	
	Here, $\hat{v}$ and $\hat{v}'$ are the predicted values before and after each compression decision, produced by the critic head. While $\overline{c}$ is generated autoregressively, the remaining computations—including both losses—are fully parallelized across the trajectory using the transformer architecture.

	\needspace{5\baselineskip} % not to put the section heading at the bottom unless there's room for 5 lines.
	\section{Experiments}
	%\added[id=mahdi]
	{In this section, we detail the experimental setup used to evaluate our model’s compression capabilities. We begin by describing the dataset, followed by the training methodology and hyper-parameter choices. Finally, we outline the hardware constraints and training procedures, including specific optimization strategies employed to stabilize and enhance learning. These components collectively form the foundation for reproducible and interpretable results.}
	
	\subsection{Dataset}
	% \Arashnote{If we want to consider other datasets or just only one, you can talk about it, about the distribution of samples, max length, where you downloaded it, etc.}
	
	We utilized the enwik8 dataset, a well-recognized benchmark for compression tasks which is also used for the Hutter Prize~\cite{hutter2006prize}. This dataset includes the first 100 million bytes of the English Wikipedia dump from March 3, 2006. It can be downloaded at~\cite{mahoney2006enwik8}. As the name suggests, the dataset encompasses a wide variety of domains,%\Seyed{If you have statistics such as the number of words, sentences, document names, and the number of documents in each domain, along with other relevant details, it would be helpful to present them in a table for a clearer and more structured description of the dataset.}, 
	ensuring that our model is trained on a broad spectrum of information. 
	%\Arash{Does this ok in comparison? I mean, are the other research works did the same? and are we allowed to do the comparison? }
	
	\subsection{Hyper-parameters and Training Details}
	
	During the pre-training phase, the compressor network was trained to reproduce the input sequence exactly. This step allowed the model to develop a foundational understanding of the data distribution before transitioning to the compression task. We then fine-tuned a \gls*{t5}-small model, which contains 60 million parameters, using a standard supervised learning approach. 
	%\added[id=ghazal]
	{The choice of T5-small as our backbone model was driven by its balance between computational efficiency and functional capability. Although larger T5 variants (Base, Large, or XL) offer better pattern recognition, their memory requirements and inference latency make them impractical for consumer-level hardware.}
	
	For optimization, we employed the Adam optimizer~\cite{kingma2014adam} with a learning rate of $1 \times 10^{-6}$. The batch size was set to 16, and the maximum sequence length was limited to 128 tokens. Our implementation was built using the PyTorch framework\footnote{https://pytorch.org/} and the HuggingFace Transformers library~\cite{wolf-etal-2020-transformers}.
	
	\subsection{Hardware and Training Setup}
	
	All experiments were conducted on an NVIDIA K80 GPU with 20GB of VRAM. The model was trained for 2 days using this hardware configuration. Due to memory limitations, the batch size was restricted to 16, and the maximum sequence length was capped at 128 tokens. These constraints primarily influenced training time and computational efficiency, as larger batch sizes and longer sequence contexts could not be explored. Future work on higher-memory GPUs could investigate the impact of larger batch sizes and extended context lengths on compression performance.
	
	A key training parameter affecting performance was the token cost. According to Shannon's entropy formula, the minimum number of bits required to encode a token is $ \log_2(|V|) $, where $V$ represents the vocabulary size. To stabilize training and improve performance, we scheduled the token cost to increase gradually throughout training. Additionally, we applied reward scaling to reduce variance in policy gradient estimations, enhancing the stability of the learning process.
	
	% \Mahdidone{I recommend introducing the token cost purely as a training parameter that helps the model to converge to an optimal (if any) solution. But there is an optimal value for the token cost which is equal to log2(vocab-size).}{Arash}{That's ok. We can say that, and do some experiments and declare the optimal token cost found by mathematical procedure (we can write any equations and proofs too). After seeing the experimental results, we deduce that the optimal value is the best. However, I think if we don't do any experiments it would be ok too.}
	
	\section{Results and Analysis}
	Our method performs competitively with traditional compression algorithms while introducing a novel sequence-to-sequence modeling approach. Notably, we made no modifications to \gls*{t5}’s architecture beyond adding a specialized head for value prediction.
	
	We present results obtained on the enwik8 dataset, comparing our approach with existing methodologies. A comprehensive analysis of the algorithm and framework highlights key performance aspects. %\Seyed{Which results belong to our method? It would be better to highlight the best-performing method in bold and refer to our approach as the 'proposed method.' Additionally, among the abbreviations in the table, only \gls*{nncp} is mentioned in the text, while the other three are introduced solely in the table without any explanation before or after. It would be beneficial to provide a brief introduction to these abbreviations before Table 1. Furthermore, you should describe Table 1—are our results good or bad? Why? Provide an analysis to clarify our performance compared to other methods.}

	Our proposed method demonstrates a notable improvement over traditional dictionary-based compressors such as   \textbf{XZ} and \textbf{GZIP}, achieving a compression ratio of 4.12 compared to 4.0 for \textbf{XZ} and 2.7 for \textbf{GZIP}. However, it falls short of the state-of-the-art neural compression method \gls*{nncp}, which achieves a ratio of 6.7 (see Table~\ref{tab:results-comparison}).

	\begin{table}[ht]
		\centering
		\caption{Compression ratios for the enwik8 dataset.}
		\label{tab:results-comparison}
		\begin{tabular}{lcc}
			\toprule
			Program & Compressed Bytes & Compression Ratio \\
			\midrule
			\textbf{NNCP} & 14,915,298 & \textbf{6.7} \\
			Proposed method & 24,141,013 & 4.12 \\ 
			XZ & 24,865,244 & 4.0 \\ 
			GZIP & 36,445,248 & 2.7 \\
			\bottomrule
		\end{tabular}
	\end{table}
	
	Our method's advantage over \textbf{XZ} and \textbf{GZIP} stems from its ability to adaptively learn a compact tokenized representation, rather than relying on fixed statistical or dictionary-based models. Unlike {\gls*{nncp}, which utilizes a more computationally intensive transformer-based sequence model, our approach aims to balance efficiency with feasibility for deployment on resource-limited hardware.
		
		\subsection{Comparison with Traditional Methods}
		
		Traditional compression methods such as \textbf{XZ} and \textbf{GZIP} rely on well-established techniques:
		
		\begin{itemize}
			\item \textbf{XZ}: Utilizes the LZMA2 algorithm, an improved variant of Lempel-Ziv-Markov chain compression. It achieves high compression ratios on large datasets but at the cost of higher computational complexity.
			\item \textbf{GZIP}: Implements the DEFLATE algorithm, which combines LZ77 for dictionary-based compression with Huffman coding for entropy reduction. It is computationally efficient but achieves lower compression ratios compared to   \textbf{XZ} and learning-based methods.
		\end{itemize}
		
		Although our method does not surpass \gls*{nncp}, it provides a trade-off between compression efficiency and computational feasibility. %\added[id=ghazal]
		{Unlike \gls*{nncp}, which relies on transformers to model data dependencies, our framework uses \gls*{rl} to actively learn optimal encoding strategies based on reward feedback to be more flexible in handling diverse data types. This \gls*{rl}-driven adaptation is useful for non-stationary data distributions, where static transformer models may fail}. Further optimizations, such as increased context length and improved attention mechanisms, could help close this gap in future work.

		% \Arash{Is there other weaker methods to be added here for comparison, showcasing that our system is better than many systems?}

		During our experiments, we identified several areas for improvement that were not fully explored due to hardware limitations. One key constraint was the \textbf{compression chunk size}, which determines the number of tokens compressed independently within a group. Since our model's context length is currently limited to \textbf{128 tokens}, larger chunk sizes could potentially yield better compression ratios by leveraging more context.
		The resulting compression ratio comparison across different chunk sizes is illustrated in Figure~\ref{compression_ratio}.
		
		\begin{figure} [H]
			\centering
			\includegraphics[width=0.55\linewidth]{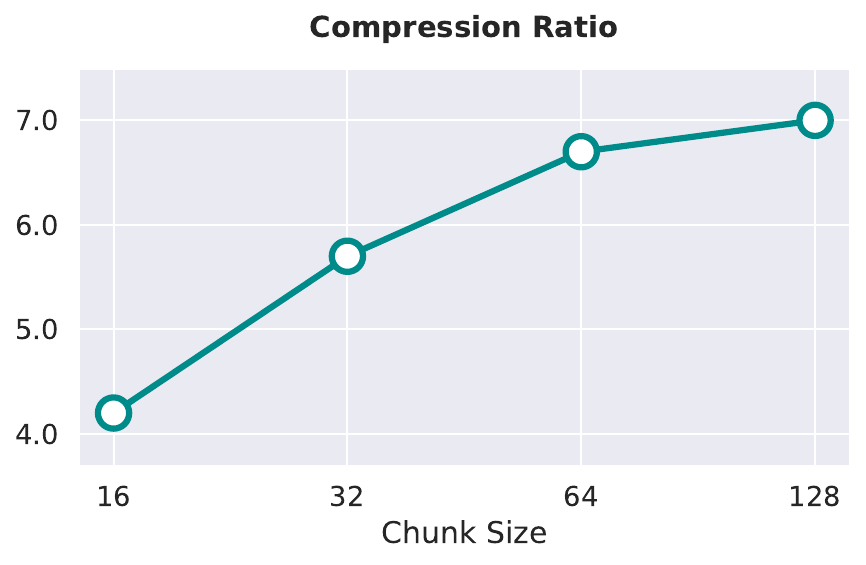}
			\caption{Compression ratio comparison across different chunk sizes.}
			\label{compression_ratio}
		\end{figure}	
		
		However, as shown in Figure~\ref{compression_latency}, processing latency is minimized at \textbf{64 tokens per chunk}, with larger chunk sizes leading to increased processing time. This trade-off suggests an optimal balance between context size and computational efficiency for real-time applications.
		
		\begin{figure} [H]
			\centering
			\includegraphics[width=0.55\linewidth]{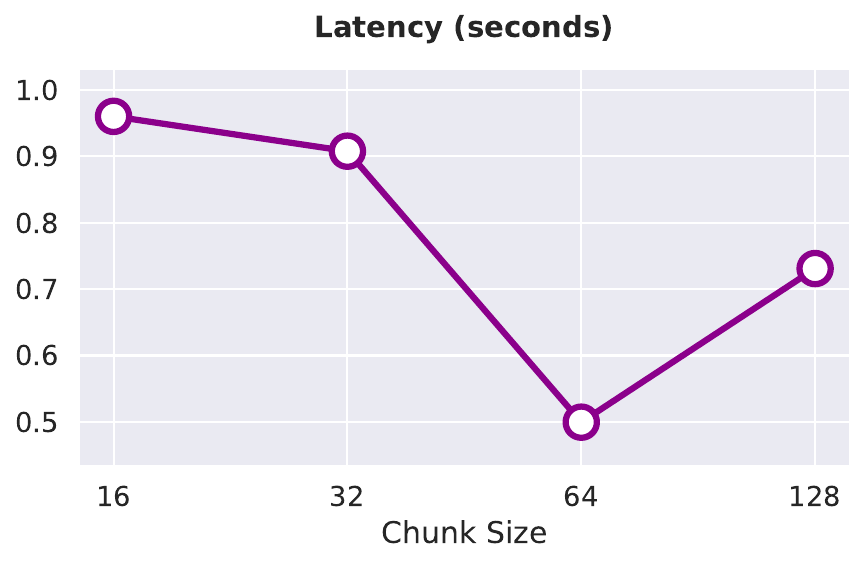}
			\caption{Compression latency vs. chunk size. Time is measured for a single batch.}
			\label{compression_latency}
		\end{figure}

		Additionally, alternative attention mechanisms could enhance compression performance by reducing redundancy more effectively. Future work could explore \textbf{adaptive chunking strategies} or \textbf{memory-efficient attention formulations} to further improve our approach.

		Since each chunk is processed independently, our method enables higher throughput by leveraging parallelism across multiple GPU cores. However, memory usage scales \textbf{quadratically with context size} and \textbf{linearly with batch size}, introducing a bottleneck as context size increases. For \textbf{non-batched compression}, throughput improves with larger chunk sizes, but our experiments were limited to \textbf{128 tokens} due to hardware constraints. 
		
		For \textbf{batched compression}, parallelism significantly improves processing speed. As shown in Figure \ref{compression_throughput}, the optimal chunk size is 64 tokens, achieving a peak throughput of \textbf{511.0 tokens per second}.
		
		\begin{figure} [H]
			\centering
			\includegraphics[width=0.55\linewidth]{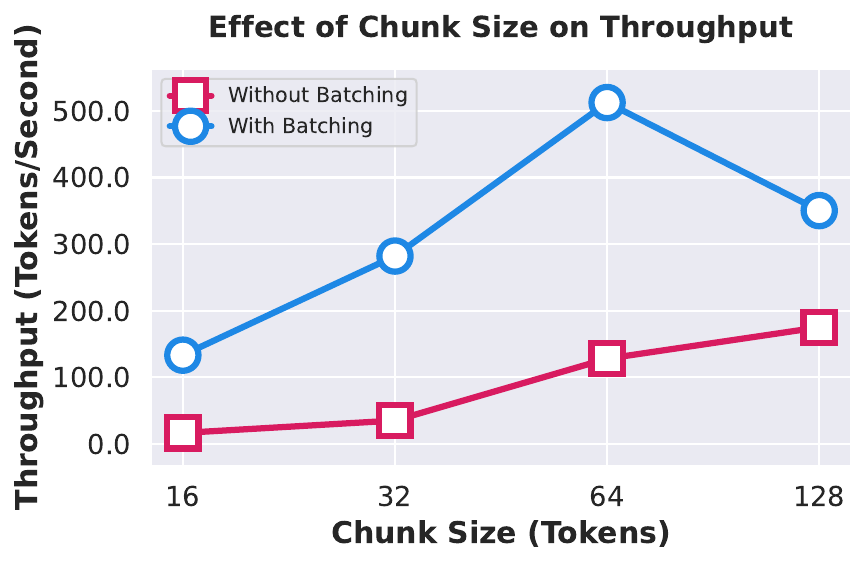}
			\caption{Compression throughput vs. chunk size.}
			\label{compression_throughput}
		\end{figure}
		
		% \Arashnote{Please insert plots and charts as many as you can regarding performance on train, dev and test sets, decoding, etc.}
		
		\section{Discussion}
		
		In this paper, we proposed a novel data compression framework that leverages \gls*{rl} to optimize compression strategies. Our method integrates the \gls*{t5} model with \gls*{rl} and demonstrates competitive performance compared to traditional lossless compression techniques. The primary advantage of our approach is the use of sequence-to-sequence (seq2seq) modeling, which allows for dynamic adjustment of the compression strategy based on the input data, rather than relying on fixed models or predefined rules. This results in a more flexible and efficient compression process, particularly when applied to diverse datasets.
		
		One of the key contributions of this work is the use of \glspl*{llm} to leverage contextual features in the data, which enables better representation and encoding of the input. Unlike auto-encoders, which typically compress data into dense, fixed-size vector representations, our approach maintains a token-based structure. This allows for a more efficient encoding, as the tokenization directly corresponds to the original data format. By preserving the token-based structure, our method better maintains the semantic integrity of the data and leads to improved compression ratios, all while avoiding the limitations of dense vector representations that are less adaptable to the diverse nature of the data.
		
		Moreover, by adopting a seq2seq approach, where the compressor and decompressor are separate models, we introduce a level of flexibility that auto-encoders lack. The ability to decouple these components means that the compressor and decompressor can be shipped separately and used independently, which can be particularly useful in real-world applications where computational resources vary. This separation allows for modular deployment, where, for example, the compressor can be used in resource-constrained environments, while the decompressor, potentially requiring more computational power, can be deployed in environments with greater hardware availability.
		
		While the approach offers significant improvements, several challenges remain. The \gls*{rl}-based optimization of sequence length requires substantial computational resources during training, making it less suitable for real-time applications on limited hardware~\cite{9387144}. However, the ability to run efficiently on personal computers suggests that our method could still be beneficial in resource-constrained environments, especially for applications where shipping the compressor separately to lightweight devices is desirable. The trade-off between compression efficiency and computational cost remains an important area for future exploration.
		
		The performance of the method is also heavily influenced by the context size and chunking strategy. As noted in the results section, larger chunk sizes tend to improve the compression ratio but introduce higher latency and reduced throughput. In contrast, smaller chunks allow for faster processing but may lead to less efficient compression. Future work could investigate adaptive chunking strategies or explore memory-efficient attention mechanisms that could further improve the balance between compression performance and computational resources.
		
		Furthermore, although the method performs well across various datasets, its generalizability to other types of data, such as images or videos, remains to be explored. Extending the approach to multimodal data compression, leveraging the inherent strengths of \glspl*{llm} for handling diverse inputs, could be a promising direction for future research.
		
		Overall, our results suggest that sequence-to-sequence modeling combined with \gls*{rl} offers a scalable and efficient solution for data compression. The separability of the compressor and decompressor also opens new possibilities for modular deployment across different environments. Future work will focus on refining the scalability of the model, improving its ability to handle large datasets, and optimizing its computational efficiency for broader real-world applications.
		
		% \Arashnote{Discuss the advantages of your proposed method. Use evidence to sohw the superiority of the proposed algorithm.}
		\section{Conclusion and Future Work}
		%\added[id=ghazal]
		{In this article, we introduced a novel, lightweight \gls*{rl}-based framework for lossless data compression, specifically designed to operate efficiently on resource-limited hardware. Our approach leverages \gls*{rl} to dynamically optimize compression policies in real time and a tokenized intermediate representation that enables more efficient and interpretable compression. These changes in transformer-based compression demonstrates considerable promise in advancing the efficiency of compression technologies and established a strong foundation for further exploration. The key novelty of our method lies in its use of \gls*{rl} to model compression as a decision-making task, where an agent learns to select context-aware encoding strategies through reward-driven exploration. This \gls*{rl} formulation not only eliminates the need for domain-specific assumptions but also enables efficient adaptation to diverse data types without excessive memory or compute demands.}%\Seyed{It is uncommon to use bullet points in the conclusion. Please remove them and incorporate their content into regular sentences.}
		
		In future work, we plan to implement an unlimited attention mechanism to improve the model's capacity to capture long-range dependencies in data. Additionally, we aim to integrate \gls*{rnn} with the Transformer architecture to enhance performance. Applying quantization techniques will help optimize resource utilization, enabling the use of larger models. Furthermore, granting the compressor access to the live state of the decompressor could lead to more informed compression decisions. Lastly, we intend to adopt relative positional embeddings to improve the model's understanding of data sequences.

		%% end main code

		%% The Appendices part is started with the command \appendix;
		%% appendix sections are then done as normal sections
		% \appendix

		% Appendix text.
		
		%% If you have bib database file and want bibtex to generate the
		%% bibitems, please use
		%%
		%%  \bibliographystyle{elsarticle-num} 
		%%  \bibliography{<your bibdatabase>}
		
		%% else use the following coding to input the bibitems directly in the
		%% TeX file.
	
	%%%
	
	% Entries for the entire Anthology, followed by custom entries
	\bibliographystyle{unsrt}  
	
	\bibliography{custom}

\end{document}